\documentclass[10pt,twocolumn,letterpaper]{article}

\usepackage{iccv}
\usepackage{times}
\usepackage{epsfig}
\usepackage{graphicx}

\usepackage{mathtools}
\usepackage{amsmath}
\usepackage{amssymb}
\usepackage{bm}

\usepackage{float}
\usepackage{subcaption}
\usepackage{wrapfig}
\usepackage[export]{adjustbox}

\usepackage{booktabs}
\usepackage{multirow}
\usepackage[table,xcdraw]{xcolor}
\usepackage{multicol}

\DeclareMathOperator*{\argmin}{arg\,min}

\usepackage[pagebackref=true,breaklinks=true,letterpaper=true,colorlinks,bookmarks=false]{hyperref}

\iccvfinalcopy 

\def\iccvPaperID{****} 

\ificcvfinal\fi

\begin{document}

\title{Neural 3D Morphable Models: Spiral Convolutional Networks for 3D Shape Representation Learning and Generation}

\author{Giorgos Bouritsas \thanks{Equal Contribution}\textsuperscript{\, \,1}
\qquad Sergiy Bokhnyak\textsuperscript{ * 2} \qquad Stylianos Ploumpis\textsuperscript{1,3}\\ \qquad Michael Bronstein\textsuperscript{1,2,4} \qquad Stefanos Zafeiriou\textsuperscript{1,3}\\
Imperial College London, UK\textsuperscript{1} \quad Universita Svizzera Italiana, Switzerland\textsuperscript{2} \quad  FaceSoft.io\textsuperscript{3}  \quad  Twitter\textsuperscript{4} \\
\textsuperscript{1}{\tt\small \{g.bouritsas18, s.ploumpis, m.bronstein, s.zafeiriou\}@imperial.ac.uk}  \quad \textsuperscript{2}{\tt\small bokhns@usi.ch}}

\maketitle
\ificcvfinal\thispagestyle{empty}\fi

\begin{figure*}[h!]
\includegraphics[width=\linewidth]{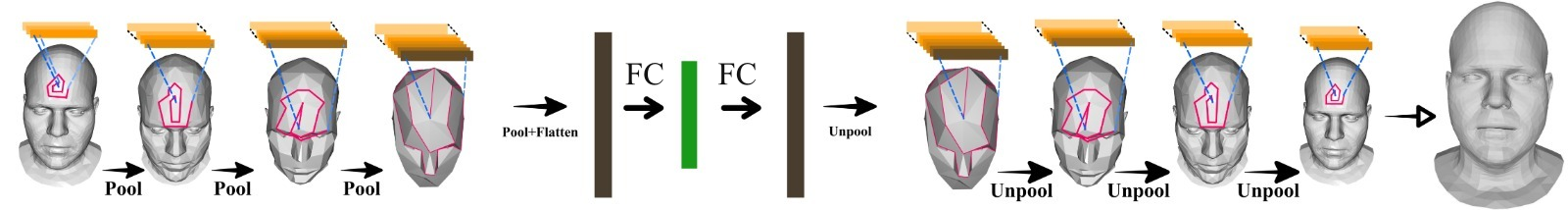}
   \caption{Illustration of our Neural3DMM architecture}
\label{fig:architecture}
\end{figure*}

\begin{abstract}
Generative models for 3D geometric data arise in many important applications in 3D computer vision and graphics. In this paper, we focus on 3D deformable shapes that share a common topological structure, such as human faces and bodies. Morphable Models and their variants, despite their linear formulation,  have been widely used for shape representation, while most of the recently proposed non-linear approaches resort to intermediate representations, such as 3D voxel grids or 2D views. 
In this work, we introduce a novel graph convolutional operator, acting directly on the 3D mesh, that explicitly models the inductive bias of the fixed underlying graph. This is achieved by enforcing consistent local orderings of the vertices of the graph, through the spiral operator, thus breaking the permutation invariance property that is adopted by all the prior work on Graph Neural Networks. 
Our operator comes by construction with desirable properties (anisotropic, topology-aware, lightweight, easy-to-optimise), and by using it as a building block for traditional deep generative architectures, we demonstrate state-of-the-art results on a variety of 3D shape datasets compared to the linear Morphable Model and other graph convolutional operators.
\end{abstract}

\section{Introduction}

The success of deep learning in computer vision and image analysis, speech recognition, and natural language processing, has driven the recent interest in developing similar models for 3D geometric data. Generalisations of successful architectures such as convolutional neural networks (CNNs) to data with non-Euclidean structure (e.g. manifolds and graphs) is known under the umbrella term {\em Geometric deep learning} \cite{bronstein2017geometric}. In applications dealing with 3D data, the key challenge of geometric deep learning is a meaningful definition of intrinsic operations analogous to convolution and pooling on meshes or point clouds. Among numerous advantages of working directly on mesh or point cloud data is the fact that it is possible to build invariance to shape transformations (both rigid and nonrigid) into the architecture, as a result allowing to use significantly simpler models and much less training data. So far, the main focus of research in the field of geometric deep learning has been on {\em analysis} tasks, encompassing shape classification and segmentation \cite{qi2017pointnet, qi2017pointnet++}, local descriptor learning, correspondence, and retrieval \cite{monti2017geometric, boscaini2016learning,masci2015geodesic}.

On the other hand, there has been limited progress in representation learning and generation of geometric data (\textit{shape synthesis}). Obtaining descriptive and compact representations of meshes and point clouds is essential for downstream tasks such as classification and 3D reconstruction, when dealing with limited labelled training data. Additionally, geometric data synthesis is pivotal in applications such as 3D printing, computer graphics and animation, virtual reality, and game design, and can heavily assist graphics designers and speed-up production. Furthermore, given the high cost and time of acquiring quality 3D data, geometric generative models can be used as a cheap alternative for producing training data for geometric ML algorithms.

Most of the previous approaches in this direction rely on intermediate representations of 3D shapes, such as point clouds \cite{achlioptas2017learning}, voxels \cite{wu2016learning} or mappings to a flat domain \cite{moschoglou20193dfacegan,ben2018multi} instead of direct surface representations, such as meshes. Despite the success of such techniques, they either suffer from high computational complexity (\eg voxels) or absence of smoothness of the data representation (\eg point clouds), while usually pre- and post-processing steps are needed in order to obtain the output surface model. Learning directly on the mesh was only recently explored in \cite{litany2018deformable, coma, wang2018pixel2mesh, kolotouros2019convolutional} for shape completion, non-linear facial morphable model construction and 3D reconstruction from single images, respectively. 

In this paper, we propose a novel representation learning and generative framework for fixed topology meshes. For this purpose, we formulate an ordering-based graph convolutional operator, contrary to the permutation invariant operators in the literature of Graph Neural Networks. In particular, similarly to image convolutions, for each vertex on the mesh, we enforce an explicit ordering of its neighbours, allowing a ``1-1'' mapping between the neighbours and the parameters of a learnable local filter. The order is obtained via a spiral scan, as proposed in \cite{spirals}, hence the name of the operator, \textit{Spiral Convolution}. This way we obtain anisotropic filters without sacrificing computational complexity, while simultaneously we explicitly encode the fixed graph connectivity. The operator can potentially be generalised to other domains that accept implicit local orderings, such as arbitrary mesh topologies and point clouds, while it is naturally equivalent to traditional grid convolutions. Via this equivalence, common CNN practices, such as dilated convolutions, can be easily formulated for meshes.

We use spiral convolution as a basic building block for hierarchical intrinsic mesh autoencoders, which we coin \textit{Neural 3D Morphable Models}. We quantitatively evaluate our methods on several popular datasets: human faces with different expressions (COMA \cite{coma}) and identities (Mein3D \cite{booth2018large}) and human bodies with shape ad pose variation (DFAUST \cite{dfaust:CVPR:2017}). Our model achieves state-of-the-art reconstruction results, outperforming the widely used linear 3D Morphable Model \cite{vetterMM} and the COMA autoencoder \cite{coma}, as well other graph convolutional operators, including the initial formulation of the spiral operator \cite{spirals}. We also qualitatively assess our framework showing `shape arithmetic' in the latent space of the autoencoder and by synthesising facial identities via a spiral convolution Wasserstein GAN. 

\section{Related Work}

\noindent\textbf{Generative models for arbitrary shapes:}
Perhaps the most common approaches for generating arbitrary shapes are \textbf{volumetric CNNs} \cite{wu20153d,qi2016volumetric,maturana2015voxnet} acting on 3D voxels. For example, voxel regression from images \cite{girdhar2016learning}, denoising autoencoders  \cite{sharma2016vconv} and voxel-GANs \cite{wu2016learning} have been proposed. Among the key drawbacks of volumetric methods are their inherent high computational complexity and that they yield coarse and redundant representations. 
\textbf{Point clouds} are a simple and lightweight alternative to volumetric representation recently gaining popularity. Several methods have been proposed for representation learning of fixed-size point clouds \cite{achlioptas2017learning} using the PointNet \cite{qi2017pointnet} architecture. In \cite{yang2018foldingnet}, point clouds of arbitrary size can be synthesised via a 2D grid deformation.
Despite their compactness, point clouds are not popular for realistic and high-quality 3D geometry generation due to their lack of an underlying smooth structure. \textbf{Image-based} methods have also been proposed, such as multi-view \cite{arsalan2017synthesizing} and flat domain mappings such as UV maps \cite{moschoglou20193dfacegan, ben2018multi}, however they are computationally demanding, require pre- and post-processing steps and usually produce undesirable artefacts. It is also worth mentioning the recently introduced \textbf{implicit-surface} based approaches \cite{mescheder2018occupancy, chen2018learning, park2019deepsdf}, that can yield accurate results, though with the disadvantage of slow inference (dense sampling of the 3D space followed by marching cubes).\\
\noindent\textbf{Morphable models:}
In the case of deformable shapes, such as faces, bodies, hands \etc, where a fixed topology can be obtained by establishing dense correspondences with a template, the most popular methods are still statistical models given their simplicity.
For \textbf{Faces}, the baseline is the PCA-based 3D Morphable Model (3DMM) \cite{vetterMM}. The Large Scale Face Model (LSFM) \cite{booth2018large} was proposed for facial identity and made publicly available, \cite{cao2014facewarehouse, FLAME:SiggraphAsia2017} were proposed for facial expression, while for the entire head a large scale model was proposed in \cite{ploumpis2019combining}. For \textbf{Body \& Hand}, the most well known models are the skinned vertex-based models SMPL \cite{SMPL:2015} and MANO \cite{MANO:SIGGRAPHASIA:2017}, respectively.
SMPL and MANO are non-linear and require (a) joint localisation and (b) solving special optimisation problems in order to project a new shape to the space of the models. In this paper, we take a different approach introducing a new family of differentiable Morphable Models, which can be applied on a variety of objects, with strong (\ie body) and less strong (\ie face) articulations. Our methods  have better representational power and also do not require any additional supervision. \\
\noindent\textbf{Geometric Deep Learning }is a set of recent methods trying to generalise neural networks to non-Euclidean domains such as graphs and manifolds \cite{bronstein2017geometric}. Such methods have achieved promising results in geometry processing and computer graphics  \cite{masci2015geodesic,boscaini2016learning}, computational chemistry \cite{duvenaud2015convolutional,gilmer2017neural}, and network science  \cite{kipfGCN,monti2017geometric}. Multiple approaches have been proposed to construct convolution-like operations, including spectral methods \cite{bruna2013spectral,defferrard2016convolutional,kipfGCN,yi2017syncspeccnn}, local charting based \cite{masci2015geodesic, boscaini2016learning, monti2017geometric, fey2018splinecnn, spirals} and soft attention \cite{velivckovic2017graph,verma2018feastnet}. Finally, graph or mesh coarsening techniques \cite{defferrard2016convolutional,ying2018hierarchical} have been proposed, equivalent to image pooling.

\section{Spiral Convolutional Networks}

\subsection{Spiral Convolution}\label{spiral_theory}

For the following discussion, we assume to be given a manifold, discretised as a triangular mesh $\mathcal{M} = (\mathcal{V,E,F})$ where $\mathcal{V}=\{1, \dots, n\}$, $\mathcal{E}$, and $\mathcal{F}$ denote the sets of vertices, edges, and faces respectively. Furthermore, let ${f:V\rightarrow\mathbb{R}}$, a function representing the vertex features.  

One of the key challenges in developing convolution-like operators on graphs or manifolds is the lack of a global system of coordinates that can be associated with each point. The first intrinsic mesh convolutional architectures such as GCNN \cite{masci2015geodesic}, ACNN \cite{boscaini2016learning} or MoNet \cite{monti2017geometric} overcame this problem by constructing a {\em local} system of coordinates $\mathbf{u}(x,y)$ around each vertex $x$ of the mesh, 
in which a set of local weighting functions $w_1,\hdots, w_L$ is applied to aggregate information from the vertices $y$ of the neighborhood $\mathcal{N}(x)$. This allows to define `patch operators' generalising the sliding window filtering in images: 
\begin{equation}\label{soft_attn}
    (f\star g)_x = \sum_{\ell=1}^L g_\ell \sum_{y\in\mathcal{N}(x)} w_\ell(\mathbf{u}(x,y)) f(y)
\end{equation}
where $\sum_{y\in\mathcal{N}(x)} w_\ell(\mathbf{u}(x,y)) f(y)$ are `soft pixels' ($L$ in total), $f$ are akin to pixel intensity in images,  and $g_\ell$ the filter weights.  The problem of the absence of a global coordinate system is equivalent to the absence of canonical ordering of the vertices, and the patch-operator based approaches can be also interpreted as attention mechanisms, as in \cite{velivckovic2017graph} and \cite{verma2018feastnet}. In particular, the absence of ordering does not allow the construction of a ``1-1'' mapping between neighbouring features $f(y)$ and and filter weights $g_\ell$, thus a ``all-to-all'' mapping is performed via learnable \textit{soft-attention} weights $w_\ell(\mathbf{u}(x,y))$. In the Euclidean setting, such operators boil down to the classical convolution, since an ordering can be obtained via the global coordinate system.

Besides the lack of a global coordinate system, another motivation for patch-operator based approaches when working on meshes, is the need for \textit{insensitivity to meshing} of the continuous surface, \ie ideally, each patch operator should be independent of the underlying graph topology.

\begin{figure}[h]
    \centering
    \begin{subfigure}{0.35\columnwidth}
        \centering
        \includegraphics[width=\linewidth]{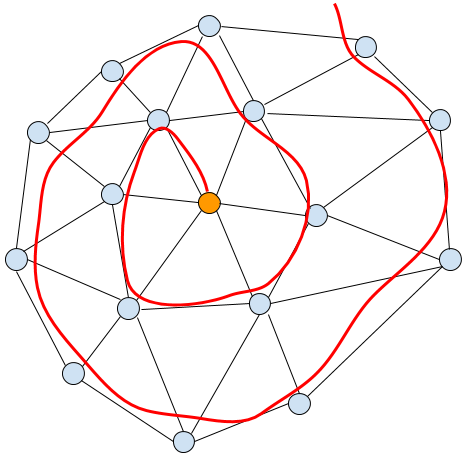}
        \label{spiral_mesh}
    \end{subfigure}
    \begin{subfigure}{0.33\columnwidth}
        \centering
        \includegraphics[width=\linewidth]{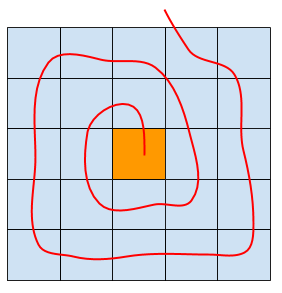}
        \label{spiral_image}
    \end{subfigure}
    \caption{Spiral ordering on a mesh and an image patch}
    \label{spiral_fig}
\end{figure}   

However, all the methods falling into this family, come at the cost of high computational complexity and parameter count and can be hard to optimise. Moreover, patch-operator based methods specifically designed for meshes, require hand-crafting and pre-computing the local systems of coordinates. To this end, in this paper we make a crucial observation in order to overcome the disadvantages of the aforementioned approaches: the issues of the absence of a global ordering and insensitivity to graph topology are irrelevant when dealing with {\em fixed topology} meshes. In particular, one can locally order the vertices and keep the order fixed. Then, graph convolution can be defined as follows:
\begin{equation}\label{ordering_based}
    (f\star g)_x = \sum_{\ell=1}^L g_\ell f(x_\ell). 
\end{equation}
where $\{x_1, \hdots, x_L\}$ denote the neighbours of vertex $x$ ordered in a fixed way. Here, in analogy with the patch operators, each patch operator is a single neighbouring vertex. 

In the Euclidean setting, the order is simply a raster scan of pixels in a patch. On meshes, we opt for a simple and intuitive ordering using spiral trajectories inspired by \cite{spirals}. Let $x\in \mathcal{V}$ be a mesh vertex, and let $R^d(x)$ be the $d$-{\em ring}, \ie an  ordered set of vertices whose shortest (graph) path to $x$ is exactly $d$ hops long; $R_j^d(x)$ denotes the $j$th element in the $d$-ring (trivially, $R_1^0(x) = x$).  
We define the {\em spiral patch operator} as the ordered sequence 
\begin{equation}
    S(x) = \{x, R_1^1(x), R_2^1(x), \hdots, R^h_{|R^h|}\},
\end{equation}
where $h$ denotes the patch radius, similar to the size of the kernel in classical CNNs. Then, {\em spiral convolution} is:
\begin{equation}
    (f * g)_x = \sum_{\ell=1}^{L} g_\ell \, f\big(S_\ell(x)\big).  
\end{equation}
The uniqueness of the ordering is given by fixing two degrees of freedom: the direction of the rings and the first vertex $R_1^1(x)$. The rest of the vertices of the spiral are ordered inductively. The direction is chosen by moving clockwise or counterclockwise, while the choice of the first vertex, the \textit{reference point}, is based on the underlying geometry of the shape to ensure the robustness of the method. In particular, we fix a reference vertex $x_0$ on a template shape and choose the initial point for each spiral to be in the direction of the shortest geodesic path to $x_0$, \ie.
\begin{equation}
   R_1^1(x) = \argmin\limits_{y \in R^1(x)}{d_\mathcal{M}(x_0,y)}, 
\end{equation}
where $d_{\mathcal{M}}$ is the geodesic distance between two vertices on the mesh $\mathcal{M}$. In order to allow for fixed-sized spirals, we choose a fixed length $L$ as a hyper-parameter and then either truncate or zero-pad each spiral depending on its size.

\noindent \textbf{Comparison to Lim \etal \cite{spirals}}: The authors choose the starting point of each spiral at random, for  every mesh sample, every vertex, and every epoch during training. This choice prevents us from explicitly encoding the fixed connectivity, since corresponding vertices in different meshes will not undergo the same transformation (as in image convolutions). Moreover, single vertices also undergo different transformations every time a new spiral is sampled. Thus, in order for the network to obtain robustness to different spiral samples, it inevitably has to become invariant to different rotations of the neighbourhoods, thus it has reduced capacity. To this end, we emphasise the need of consistent orderings across different meshes.




Moreover, in \cite{spirals}, the authors model the vertices on the spiral via a recurrent network, which has higher computational complexity, is harder to optimise and does not take advantage of the stationary properties of the 3D shape (local statistics are repeated across different patches), which are treated by our spiral kernel with weight sharing.

\begin{figure}[h]
    \centering
    \includegraphics[scale =0.07]{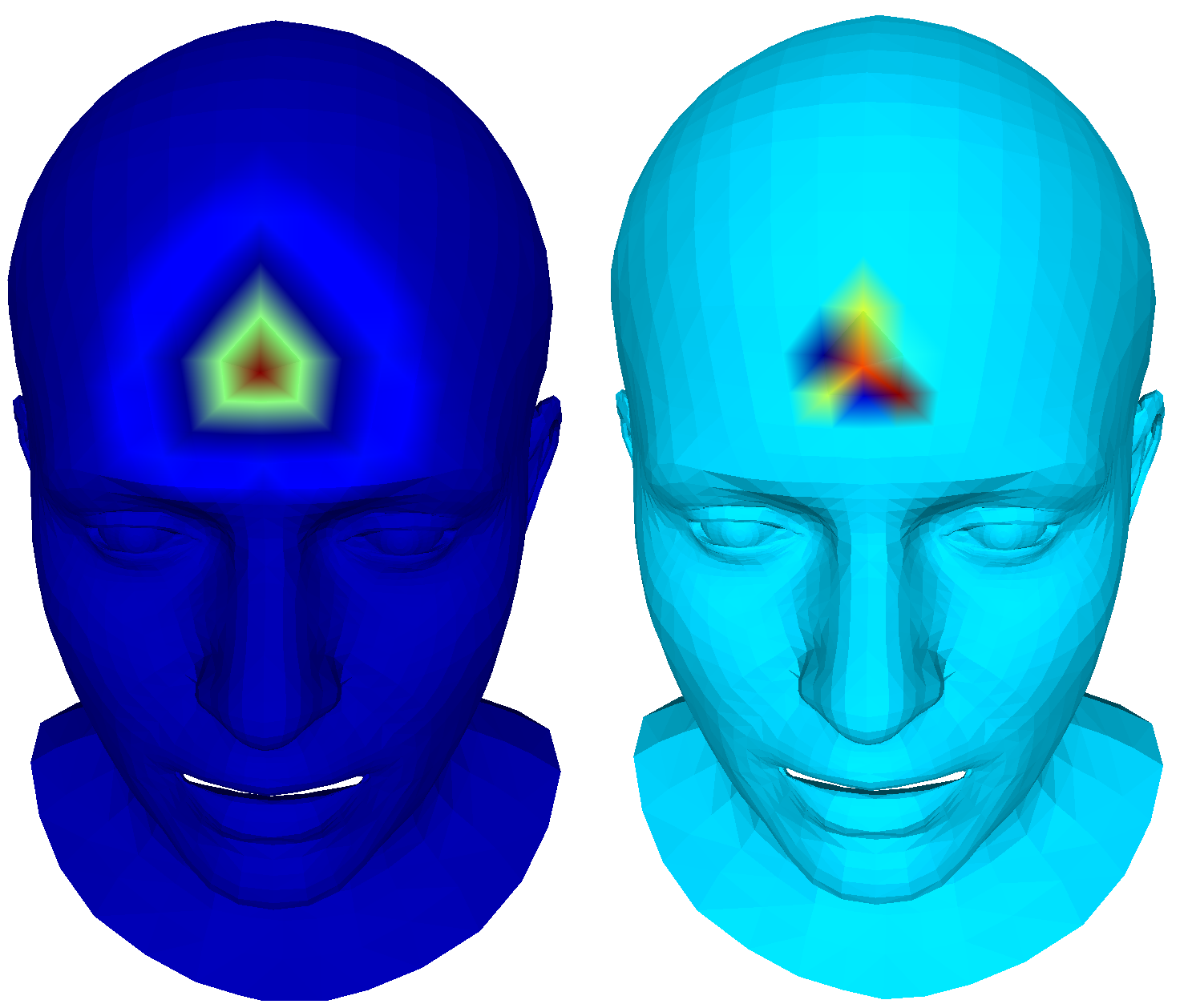}
    \caption{Activations of ChebNet vs spiral convolutions}
    \label{isotropy}
\end{figure}

\noindent \textbf{Comparison to spectral filters:} Spectral convolutional operators developed in \cite{chebnet,kipfGCN} for graphs and used in \cite{coma} for mesh autoencoders, suffer from the fact that the are inherently {\em isotropic}. This is a side-effect when one, under the absence of a canonical ordering, needs to design a permutation-invariant operator with small number of parameters. In particular, spectral filters rely on the {\em Laplacian operator}, which performs a weighted averaging of the neighbour vertices :
\begin{equation}
(\Delta f)_x = \sum\nolimits_{y: (x,y) \in \mathcal{E}} w_{xy} \big(f(y) - f(x)\big),
\end{equation}
where $w_{xy}$ denotes an edge weight. 
A polynomial of degree $r$ with learnable coefficients $\theta_0, \hdots, \theta_r$ is then applied to $\Delta$. Then, the graph convolution amounts to filtering the Laplacian eigenvalues, ${p(\Delta) = \Phi p(\Lambda) \Phi^\top}$. Equivalently:
\begin{equation}
(f * g) = p(\Delta) f = \sum_{\ell = 0}^r \theta_\ell \Delta^\ell f,
\end{equation}
While a necessary evil in general graphs, spectral filters on meshes are rather weak given that they are locally rotationally-invariant. On the other hand, spiral convolutional filters leverage the fact that on a mesh one can canonically order the neighbours. Thus, they are anisotropic by construction and as will be shown in the experimental section \ref{evaluation} they are expressive by using just one-hop neighbourhoods, contrary to the large receptive fields used in \cite{coma}. In Fig \ref{isotropy} we visualise the impulse response (centred on a vertex on the forehead) of a selected laplacian polynomial filter from the architecture of \cite{coma} (left) and from a spiral convolutional filter with $h=1$ (right). 

Finally, the equivalence of spiral convolutions to image convolutions allows the use of long-studied practices in the computer vision community. For example, small patches can be used, leading to few parameters and fast computation. Furthermore, dilated convolutions \cite{yu2015multi} can also be adapted to the spiral operator by simply sub-sampling the spiral. Finally, we argue here that our operator could be applied to other domains, such as point clouds, where an ordering of the data points can be enforced.


\subsection{Neural 3D Morphable Models}

Let $\bm{F}=[\bm{f_0}| \bm{f_1}|..., \bm{f_N}]$, $\bm{f_i} \in \mathbb{R}^{d*m}$ the matrix of all the signals defined on a set of meshes in dense correspondence that are sampled from a distribution $\mathcal{D}$, where $d$ the dimensionality of the signal on the mesh (vertex position, texture etc.) and $m$ the number of vertices. A linear 3D Morphable Model \cite{vetterMM} represents arbitrary instances $\mathbf{y}\in \mathcal{D}$ as a linear combination of the k largest eigenvectors of the covariance matrix of $\bm{F}$ by making a gaussianity assumption:
\begin{equation}
    \mathbf{y} \approx \Bar{\bm{f}} + \sum_{i}^k \alpha_i \sqrt{d_i} \mathbf{v_i} 
\end{equation}

\noindent where $\Bar{\bm{f}}$ the mean shape, $\mathbf{v_i}$ is the $i$th principal component, $d_i$ the respective eigenvalue and $\alpha_i$ the linear weight coefficient. Given its linear formulation, the representational power of the 3DMM is constrained by the span of the eigenvectors, while its parameters scale linearly w.r.t the number of the eigencomponents used, leading to large parametrisations for meshes of high resolution.

In contrast, in this paper, we use spiral convolutions as a building block to build a fully differentiable non-linear Morphable Model. In essence, a Neural 3D Morphable Model is a deep convolutional mesh autoencoder, that learns hierarchical representations of a shape. An illustration of the architecture can be found in Fig \ref{fig:architecture}. Leveraging the connectivity of the graph with spiral convolutional filters, we allow for local processing of each shape, while the hierarchical nature of the model allows learning in multiple scales. This way we manage to learn semantically meaningful representations and considerably reduce the number of parameters. Furthermore, we bypass the need to make assumptions about the distribution of the data.


Similar to traditional convolutional autoencoders, we make use of series of convolutional layers with small receptive fields followed by pooling and unpooling, for the encoder and the decoder respectively, where a decimated or upsampled version of the mesh is obtained each time and the features of the existing vertices are either aggregated or extrapolated. We follow \cite{coma} for the calculation of the features of the added vertices after upsampling, \ie through interpolation by weighting the nearby vertices with barycentric coordinates. The network is trained by minimising the $L_1$ norm between the input and the predicted output.

\subsection{Spiral Convolutional GAN}

In order to improve the synthesis of meshes of high resolution, thus increased detail, we extend our framework with a distribution matching scheme. In particular, we propose a mesh Wasserstein GAN \cite{pmlr-v70-arjovsky17a} with gradient penalty to enforce the Lipschitz constraint \cite{gulrajani2017improved}, that is trained to minimise the wasserstein divergence between the real distribution of the meshes and the distribution of those produced by the generator network. The generator and discriminator architectures, have the same structure as the decoder and the encoder of the Neural3DMM respectively. 
Via this framework, we obtain two additional properties that are inherently absent from the autoencoder: high frequency detail and a straightforward way to sample from the latent space. 

\section{Evaluation} \label{evaluation}

In this section, we showcase the effectiveness of our proposed method on a variety of shape datasets. We conduct a series of ablation studies in order to compare our operator to other Graph Neural Networks, by using the same autoencoder architecture.
Fist, we demonstrate the inherent higher capacity of spiral convolutions compared to ChebNet (spectral). Moreover, we discuss the advantages of our method compared to soft-attention based Graph Neural Networks, such as patch-operator based. Finally, we show the importance of the consistency of the ordering by comparing our method to different variants of the method proposed in \cite{spirals}.

Furthermore, we quantitatively show that our method can yield better representations than the linear 3DMM and COMA, while maintaining a small parameter count and frequently allowing a more compact latent representation. Moreover, we proceed with a qualitative evaluation of our method
by generating novel examples through vector space arithmetic. Finally, we assess our intrinsic GAN in terms of its ability to produce high resolution realistic examples. 

For all the cases, we choose as signal on the mesh the normalised deformations from the mean shape, \ie for every vertex we subtract its mean position and divide with the standard deviation. In this way, we encourage signal stationarity, thus facilitating optimisation.  The code is available at \url{https://github.com/gbouritsas/neural3DMM}.

\subsection{Datasets}
\noindent \textbf{COMA. } 
The facial expression dataset from Ranjan \etal \cite{coma}, consisting of 20K+ 3D scans (5023 vertices) of twelve unique identities performing twelve types of extreme facial expressions. We used the same data split as in \cite{coma}.

\noindent \textbf{DFAUST. }
The dynamic human body shape dataset from Bogo \etal \cite{dfaust:CVPR:2017}, consisting of 40K+ 3D scans (6890 vertices)  of ten unique identities performing actions such as leg and arm raises, jumps, etc. 
We randomly split the data into a test set of 5000, 500 validation, and 34,5K+ train.

\noindent \textbf{MeIn3D. } 
The 3D large scale facial identity dataset from Booth \etal \cite{booth20163d}, consisting of more than 10,000 distinct identity scans with 28K vertices which cover a wide range of gender
ethnicity and age. For the subsequent experiments, the MeIn3D dataset was randomly split within demographic constraints to ensure gender, ethnic and age diversity, into 9K train and 1K test meshes.

\begin{figure*}[t]
\centering
    \includegraphics[width=0.33\linewidth]{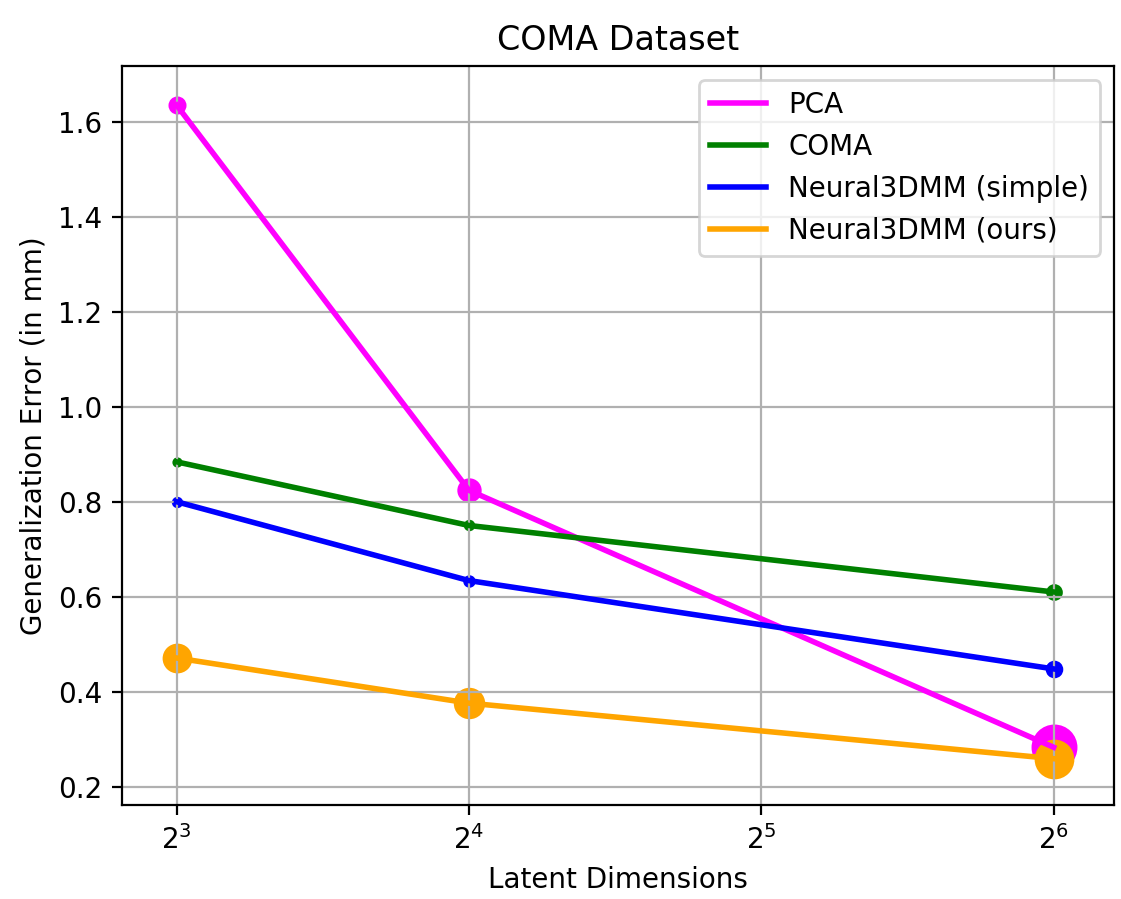}
    \includegraphics[width=0.33\linewidth]{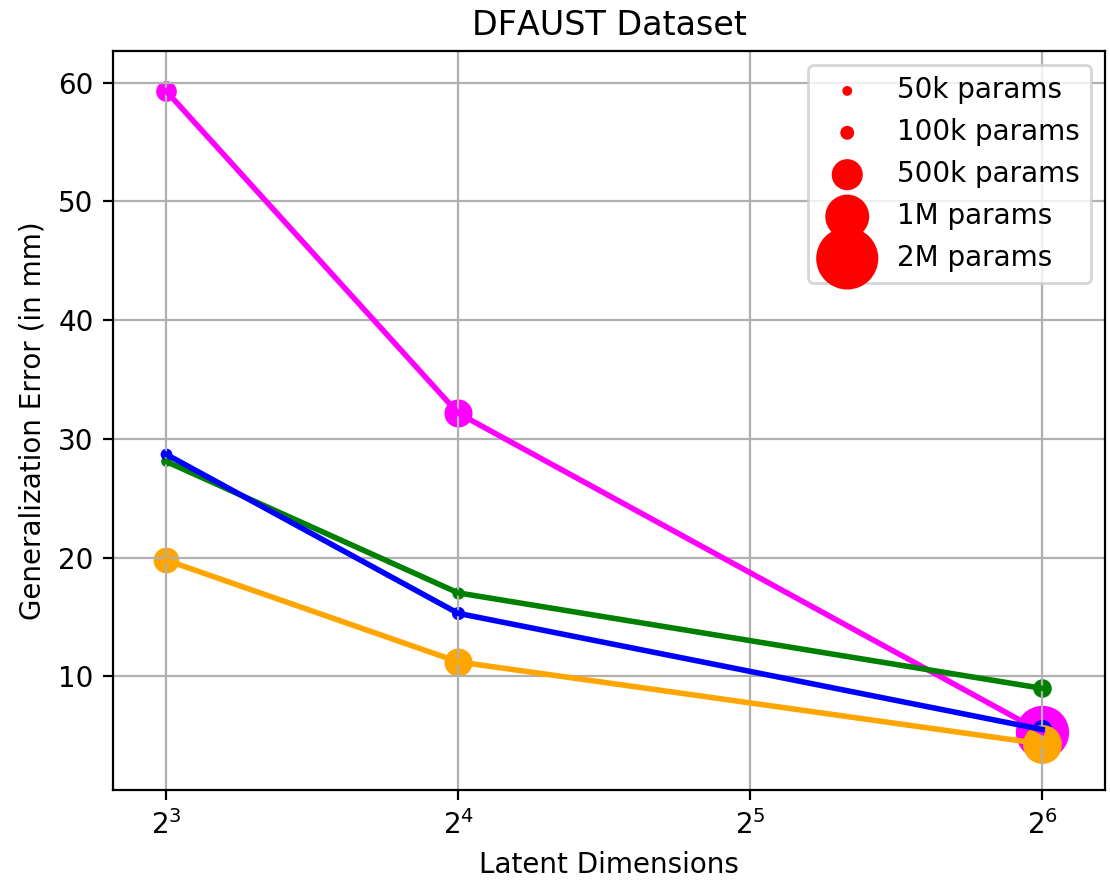}
    \includegraphics[width=0.33\linewidth]{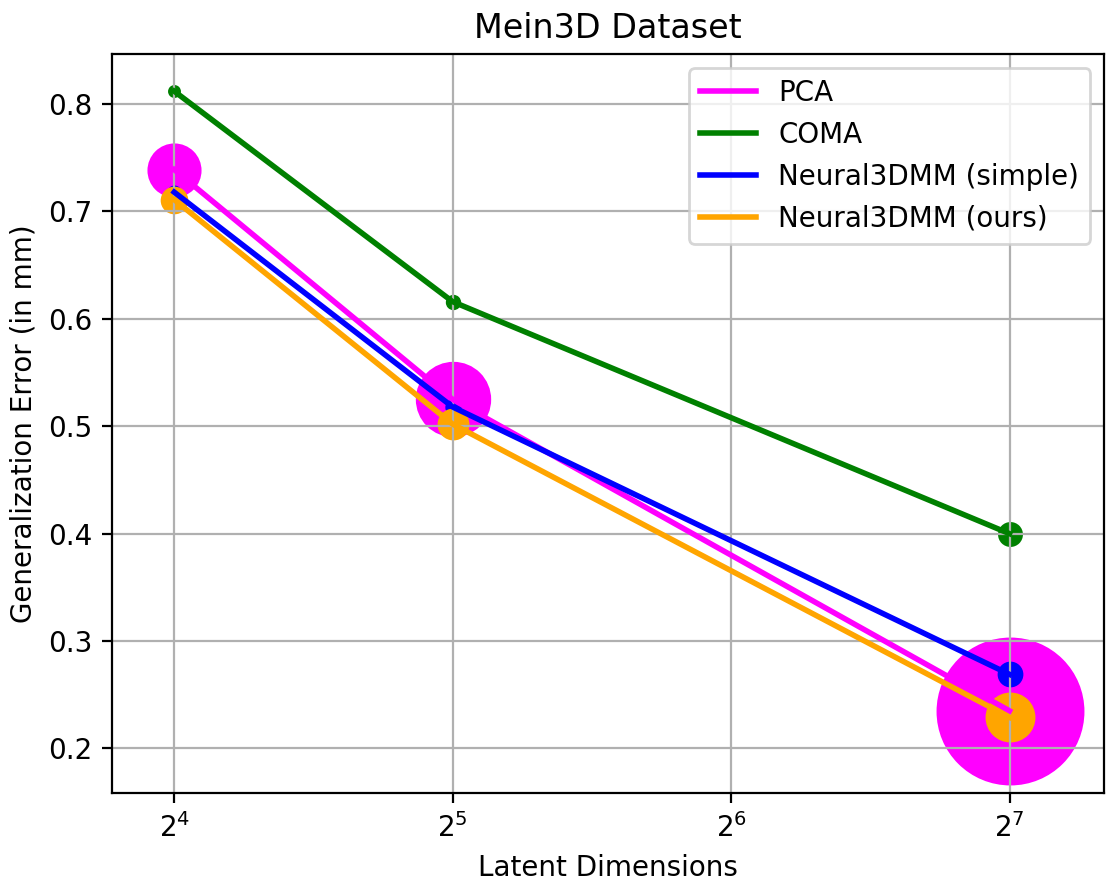}
    \caption{Quantitative evaluation of our Neural3DMM against the baselines, in terms of generalisation and \# of parameters}
    \label{quantitative}
\end{figure*}



For the quantitative experiments of sections \ref{ablation} and \ref{3dmm_comparison} the evaluation metric used is \textbf{generalisation}, which measures the ability of a model to represent novel shapes from the same distribution as it was trained on. More specifically we evaluate the average per sample and per vertex euclidean distance in the 3D space (in millimetres) between corresponding vertices in the input and its reconstruction. 

\subsection{Implementation Details}\label{implementation}

We denote as $SC(h,w)$ a spiral convolution of $h$ hops and $w$ filters, $DS(p)$ and $US(p)$ a downsampling and an upsampling by a factor of $p$, respectively, $FC(d)$ a fully connected layer, $l$ the number of vertices after the last downsampling layer. The simple Neural3DMM for COMA and DFAUST datasets is the following: 

$\mathit{Enc:} \, SC(1,16)\rightarrow DS(4)\rightarrow SC(1,16)\rightarrow DS(4)\rightarrow SC(1,16)\rightarrow DS(4)\rightarrow SC(1,32)\rightarrow DS(4)\rightarrow FC(d)$

$\mathit{Dec:} \,FC(l*32)\rightarrow US(4)\rightarrow SC(1,32)\rightarrow US(4)\rightarrow SC(1,16)\rightarrow US(4)\rightarrow SC(1,16)\rightarrow US(4)\rightarrow SC(1,3)$ 

For Mein3D, due to the high vertex count, we modified the COMA architecture for our simple Neural3DMM by adding an extra convolution and an extra downsampling/upsampling layer in the encoder and the decoder respectively (encoder filter sizes: [8,16,16,32,32], decoder: mirror of the encoder). The larger Neural3DMM follows the above architecture, but with an increased parameter space. For COMA, the convolutional filters of the encoder had sizes [64,64,64,128] and for Mein3D the sizes were [8,16,32,64,128], while the decoder is a mirror of the encoder. For DFAUST, the sizes were [16,32,64,128] and [128,64,32,32,16] and dilated convolutions with $h=2$ hops and dilation ratio $r=2$ were used for the first and the last two layers of the encoder and the decoder respectively. 
We observed that by adding an additional convolution at the very end (of size equal to the size of the input feature space), training was accelerated. All of our activation functions were ELUs \cite{ELU}. Our learning rate was $10^{-3}$ with a decay of $0.99$ after each epoch, and our weight decay was $5\times 10^{-5}$. All models were trained for 300 epochs. 

\begin{figure}[h]
\large
    \begin{subfigure}{\columnwidth}
        \centering
        \includegraphics[width=0.6\linewidth]{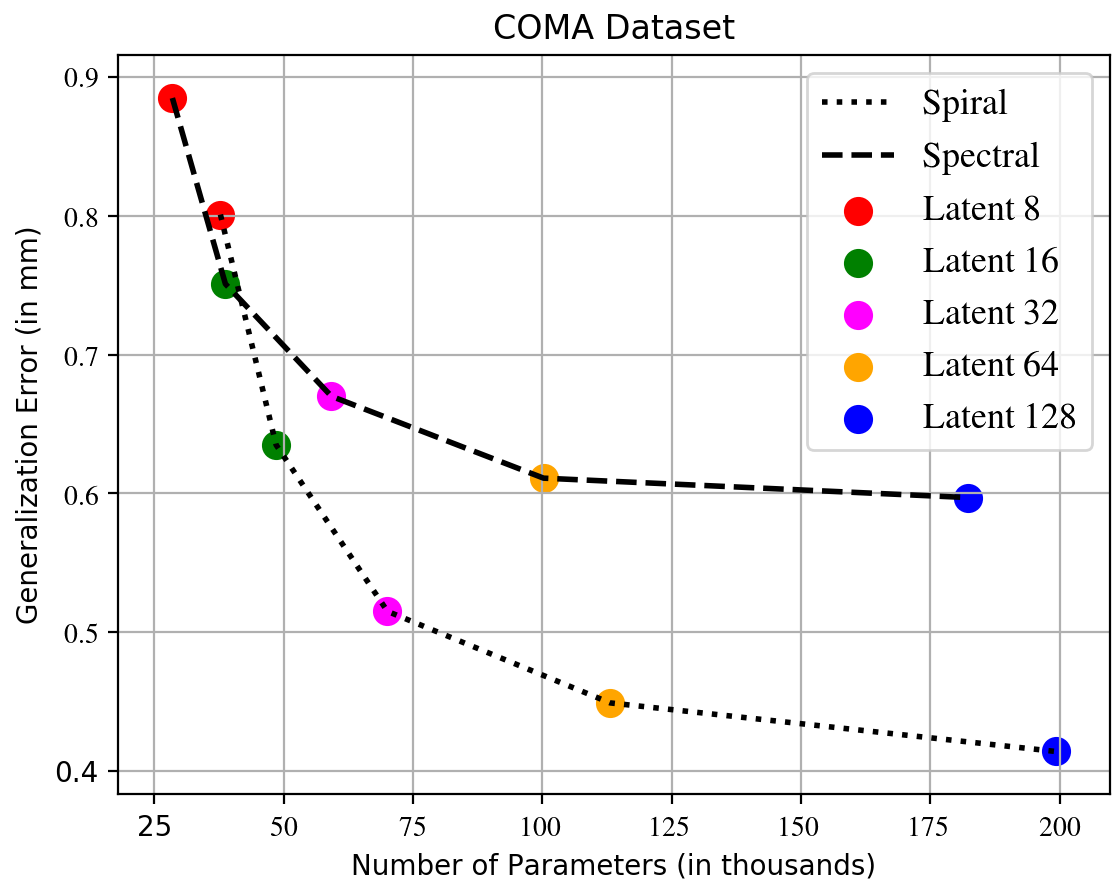}
        \label{fig:coma_tropic_comparison}
    \end{subfigure}
    \begin{subfigure}{\columnwidth}
        \centering
        \includegraphics[width=0.6\linewidth]{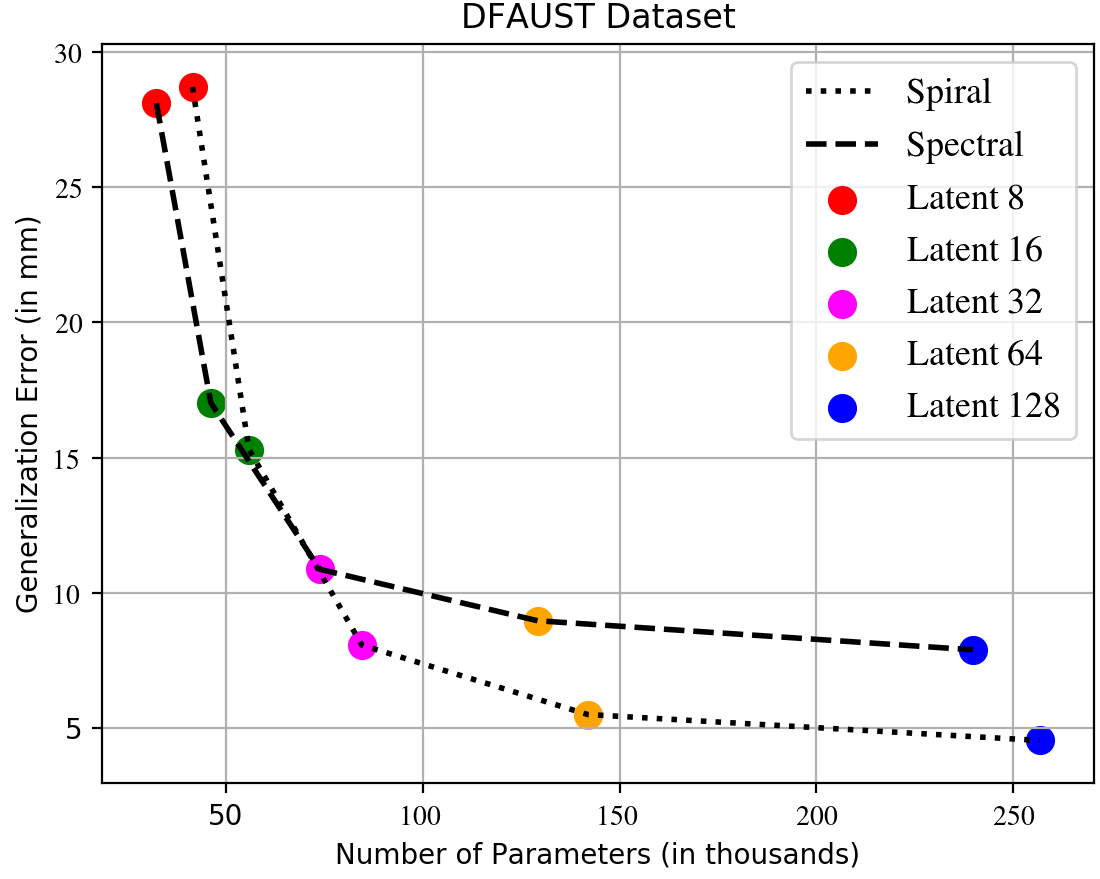}
        \label{fig:dfaust_tropic_comparison}
    \end{subfigure}
    \caption{Spiral vs ChebNet (spectral) filters}
    \label{fig:spectral_spiral_comparison}
\end{figure}

\subsection{Ablation Studies}\label{ablation}
\subsubsection{Isotropic vs Anisotropic Convolutions}\label{iso_comaparison}

For the purposes of this experiment we used the architecture deployed by the authors of \cite{coma}. The number of parameters in our case is slightly larger due to the fact that the immediate neighbours, that affect the size of the spiral, range from 7 to 10, while the polynomials used in \cite{coma} go up to the 6th power of the Laplacian. For both datasets, as clearly illustrated in Fig \ref{fig:spectral_spiral_comparison}, spiral convolution-based autoencoders consistently outperform the spectral ones for every latent dimension, in accordance with the analysis made in section \ref{spiral_theory}. 
Additionally, increasing the latent dimensions, our model's performance increases at a higher rate than its counterpart. Notice that the number of parameters scales the same way as the latent size grows, but the spiral model makes better use of the added parameters especially looking at dimensions 16, 32, 64, and 128. Especially on the COMA dataset, the spectral model seems to be flattening between 64 and 128 while the spiral is still noticeably decreasing.

\subsubsection{Spiral vs Attention based Convolutions}\label{attention_comaparison}
In this experiment we compare our method with certain state-of-the-art soft-attention based Graph Neural Networks: \textbf{MoNet}: the patch-operator based model of \cite{monti2017geometric}, where the attention weights are the learnable parameters of gaussian kernels defined on a pseudo-coordinate space\footnote{here we display the best obtained results when choosing the pseudo-coordinates to be local cartesian.}, \textbf{FeastNet} \cite{verma2018feastnet} and \textbf{Graph Attention} \cite{velivckovic2017graph}, where the attention weights are learnable functions of the input features.

In table \ref{soft_attn}, we provide results on COMA dataset, using the simple Neural3DMM architecture with latent size 16. We choose the number of attention heads (gaussian kernels in \cite{monti2017geometric}) to be either 9 (equal to the size of the spiral in our method, for a fair comparison) or 25 (as in \cite{monti2017geometric}, to showcase the effect of over-parametrisation). When it comes to similar number of parameters our method manages to outperform its counterparts, while compared to over-parametrised soft attention networks it either outperforms them, or achieves slightly worse performance. This shows that the spiral operator can make more efficient use of the available learnable parameters, thus being a lightweight alternative to attention-based methods without sacrificing performance. Also, its formulation allows for fast computation; in table \ref{soft_attn} we measure per mesh inference time in ms (on a GeForce RTX 2080 Ti GPU).

\begin{table}[h!]
\setlength{\tabcolsep}{5.3pt}
\renewcommand{\arraystretch}{1} 
\centering
\scriptsize
\begin{tabular}{|l||c|c||c|c||c|c||c|}
\hline
& \multicolumn{2}{c||}{GAT} & \multicolumn{2}{c||}{FeastNet} & \multicolumn{2}{c||}{MoNet} & Ours\\\hline
kernels & 9 & 25 & 9 &  25 &  9 &  25 & - \\\hline
error & 0,762 & 0,732 & 0,750 & 0,623 & 0,708 & \textbf{0,583}  & \textbf{\textcolor{red}{0,635}} \\\hline
params & \textcolor{red}{50K} & 101K & \textcolor{red}{49K} & 98K & \textcolor{red}{48K} & 95K& \textbf{\textcolor{red}{48K}} \\\hline
time & 12,77 & 15,37 & 9,04 & 9,66 & 10,55 & 10,96 & \textbf{\textcolor{red}{8,18}}\\\hline
\end{tabular}
\caption{Spirals vs soft-attention operators}
\label{spiral_vs_attn}
\end{table}
\subsubsection{Comparison to Lim et al. \cite{spirals}}\label{spiral_comaparison}
In order to showcase how the operator behaves when the ordering is not consistent, we perform experiments under four scenarios: the original formulation of \cite{spirals}, where each spiral is randomly oriented for every mesh and every epoch (\textit{rand mesh \& epoch)}; choosing the same orientation across all the meshes randomly at every epoch (\textit{rand epoch}); choosing different orientations for every mesh, but keeping them fixed across epochs (\textit{rand mesh}); and fixed ordering (Ours). We compare the LSTM-based approach of \cite{spirals} and our linear projection formulation (Eq \eqref{ordering_based}). The experimental setting and architecture is the same as in the previous section. The proposed approach achieves over 28\% improved performance compared to \cite{spirals}, which substantiates the benefits of passing corresponding points through the same transformations.

\begin{table}[h!]
\renewcommand{\arraystretch}{1} 
\centering
\scriptsize
\begin{tabular}{|l|c|c|c|c|}
\hline
operation &rand mesh \& epoch &rand mesh&rand epoch&fixed ordering\\\hline
LSTM & 0.888 \cite{spirals}  & 0.880 &  0,996 & 0.792\\\hline
lin. proj. & 0.829 & 0.825 & 0.951 & \textbf{0.635 (Ours)}\\\hline
\end{tabular}
\label{text}
\caption{Importance of the ordering consistency}
\end{table}
\subsection{Neural 3D Morphable models}\label{3dmm_comparison}

\begin{figure}[h]
\centering
\includegraphics[width=0.6\linewidth]{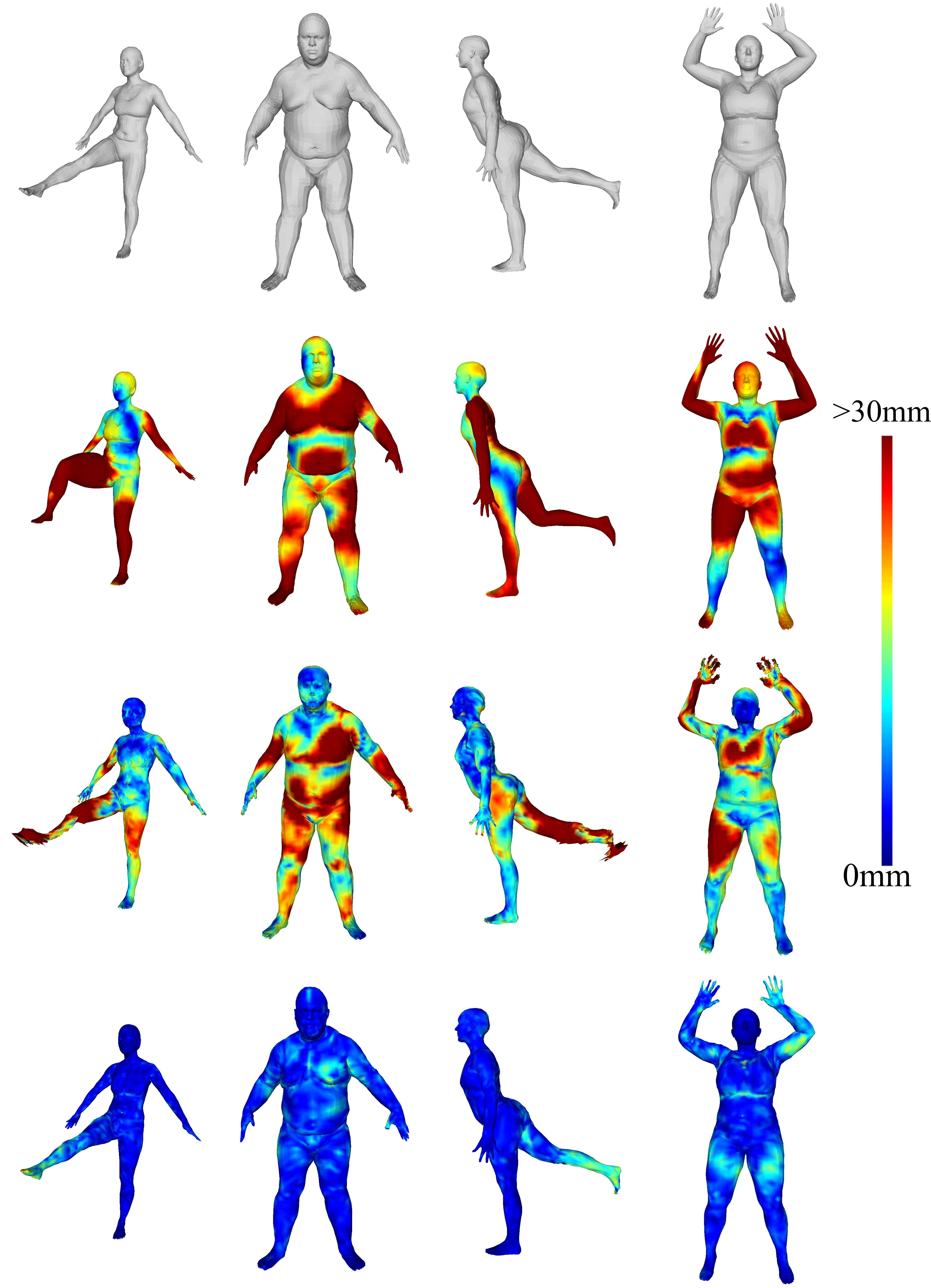}
   \caption{Colour coding of the per vertex euclidean error of the reconstructions produced by PCA (2nd), COMA (3rd), and our Neural3DMM (bottom). Top row is ground truth.}
\label{fig:dfaust_color_coding}
\end{figure}

\subsubsection{Quantitative results}

In this section, we compare the following methods for different dimensions of the latent space: \textbf{PCA}, the 3D Morphable Model \cite{vetterMM}, \textbf{COMA}, the ChebNet-based Mesh Autoencoder, \textbf{Neural3DMM (small)}, ours spiral convolution autoencoder with the same architecture as in COMA, \textbf{Neural3DMM (ours)}, our proposed Neural3DMM framework, where we enhanced our model with a larger parameter space (see Sec.~\ref{implementation}). The latent sizes were chosen based on the variance explained by PCA (explained variance of roughly 85\%, 95\% and 99\% of the total variance).

As can be seen from the graphs in Fig \ref{quantitative}, our Neural3DMM achieves smaller generalisation errors in every case it was tested on. For the COMA and DFAUST datasets all hierarchical intrinsic architectures outperform PCA for small latent sizes. That should probably be attributed to the fact that the localised filters used allow for effective reconstruction of smaller patches of the shape, such as arms and legs (for the DFAUST case), whilst PCA attempts a more global reconstruction, thus its error is distributed equally across the entire shape. This is well shown in Fig \ref{fig:dfaust_color_coding}, where we compare exemplar reconstructions of samples from the test set (latent size 16). It is clearly visible that PCA prioritises body shape over pose resulting to body parts in the wrong locations (for example see the right leg of the woman on the leftmost column). On the contrary COMA places the vertices in approximately correct locations, but struggles to recover the fine details of the shape leading to various artefacts and deformities; our model on the other hand seemingly balances these two difficult tasks resulting in quality reconstructions that preserve pose and shape. 

Comparing to \cite{coma}, it is again apparent here that our spiral-based autoencoder has increased capacity, which together with the increased parameter space, makes our larger Neural3DMM outperform the other methods by a considerably large margin in terms of both generalisation and compression. Despite the fact that for higher dimensions, PCA can explain more than 99\% of the total variance, thus making it a tough-to-beat baseline, our larger model still manages to outperform it. The main advantage here is the substantially smaller number of parameters of which we make use. This is clearly seen in the comparison for the MeIn3D dataset, where the large vertex count makes non-local methods as PCA impractical. It is necessary to mention here, that larger latent space sizes are not necessarily desirable for an autoencoder because they might lead to less semantically meaningful and discriminative representation for downstream tasks.

\subsubsection{Qualitative results}
Here, we assess the representational power of our models by the common practice of testing their ability to perform linear algebra in their latent spaces.

\begin{figure}[h]
\centering
  \includegraphics[width = 0.7\columnwidth]{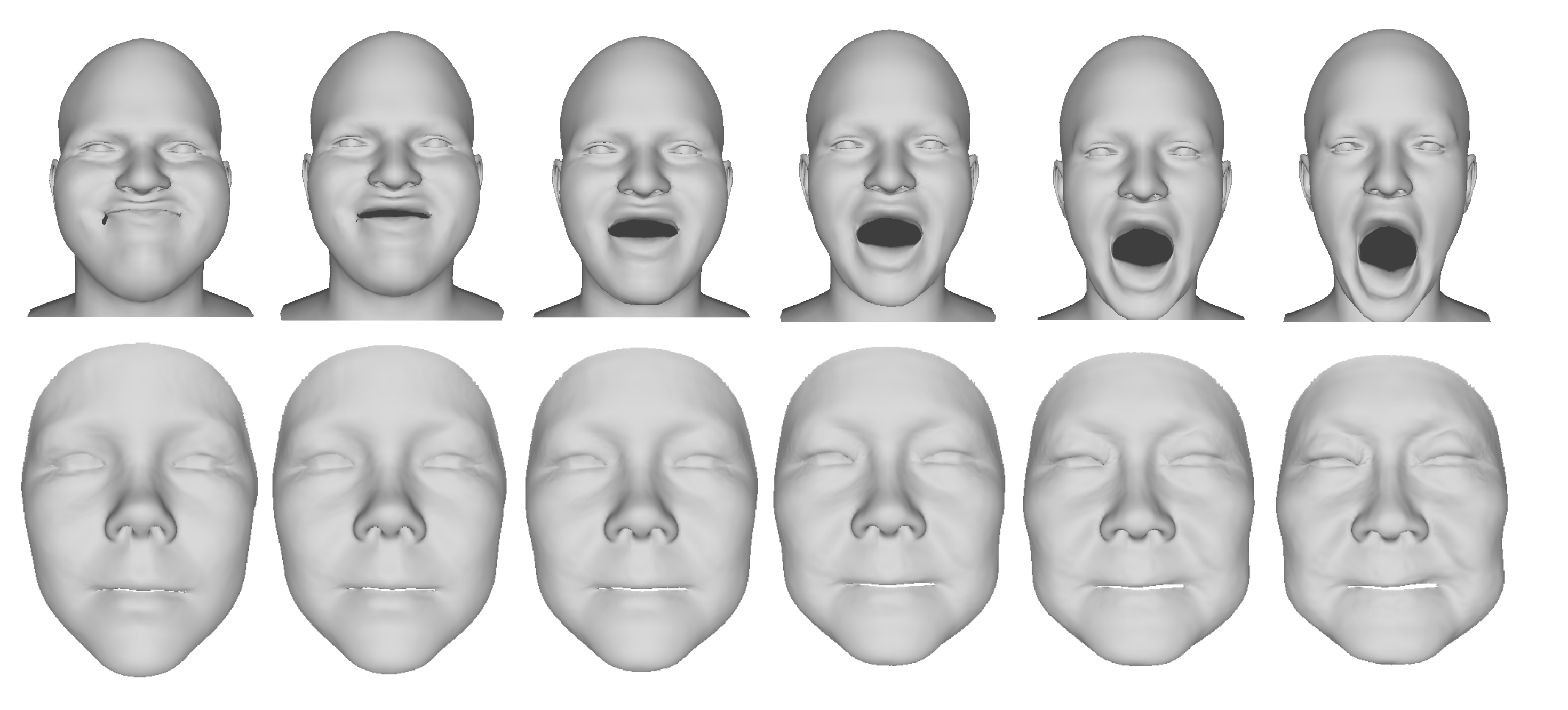}
\caption{Interpolations between expressions and identities}
\label{fig:interpolation}
\end{figure}

\noindent \textbf{Interpolation} Fig \ref{fig:interpolation}: We choose two sufficiently different samples $\mathbf{x_1}$ and $\mathbf{x_2}$ from our test set, encode them in their latent representations $\mathbf{z_1}$ and $\mathbf{z_2}$ and then produce intermediate encodings by sampling the line that connects them \ie $\mathbf{z} = a\mathbf{z_1}+(1-a)\mathbf{z_2}$, where $a\in(0,1)$.

\begin{figure}[h]
\centering
\includegraphics[width=0.6\linewidth]{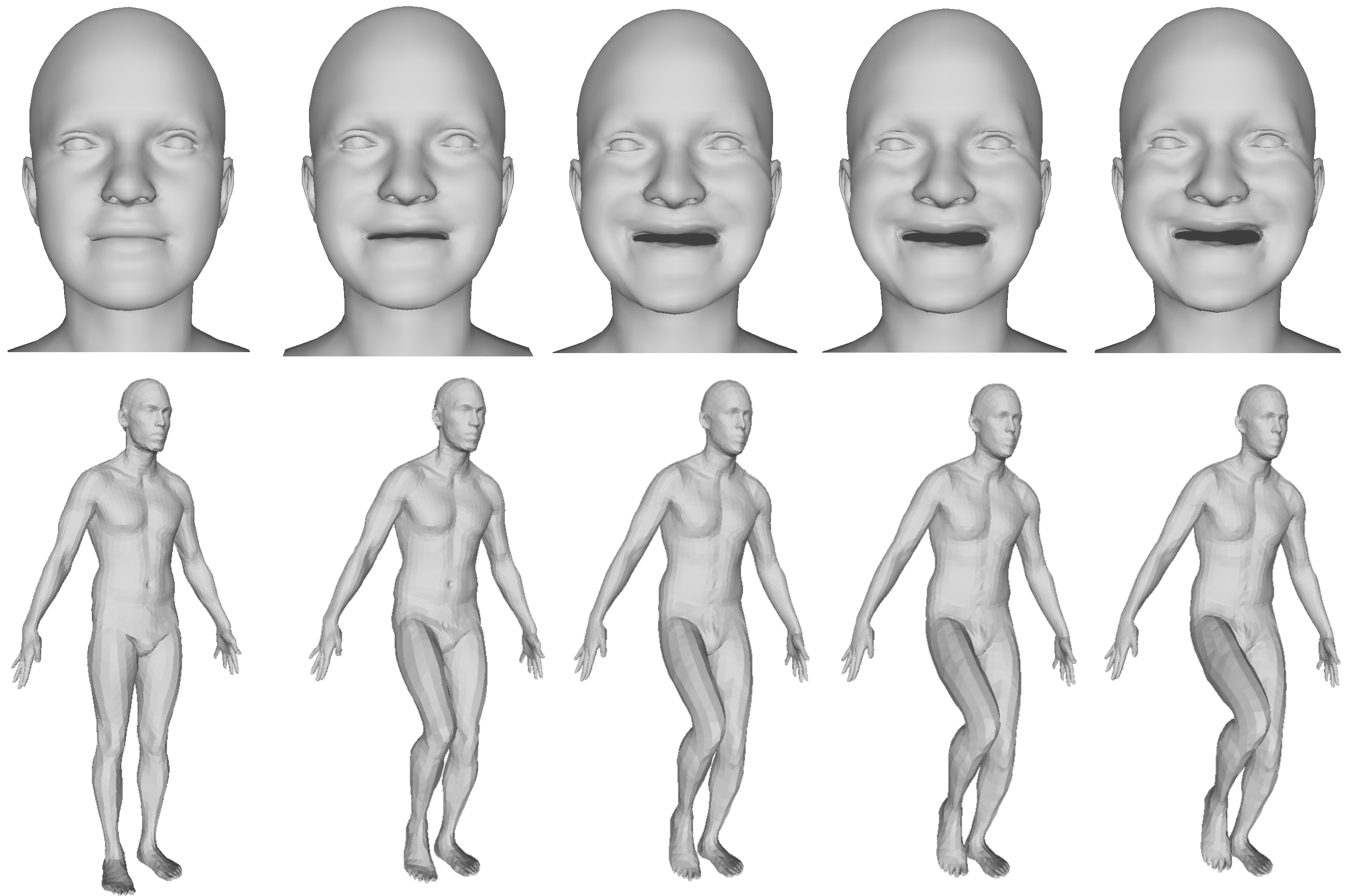}
   \caption{Extrapolation. Left: neutral expression/pose}
\label{fig:extrapolation}
\end{figure}

\noindent\textbf{Extrapolation} Fig \ref{fig:extrapolation}: Similarly, we decode latent representations that reside on the line defined by $\mathbf{z_1}$ and $\mathbf{z_2}$, but outside the respective line segment, \ie $\mathbf{z} = a*\mathbf{z_1}+(1-a)*\mathbf{z_2}$, where $a\in(-\infty,0)\cup(1,+\infty)$. We choose $\mathbf{z_1}$ to be our neutral expression for COMA and neutral pose for DFAUST, in order to showcase the exaggeration of a specific characteristic on the shape.

\begin{figure}[h]
\centering
\includegraphics[width=0.6\linewidth]{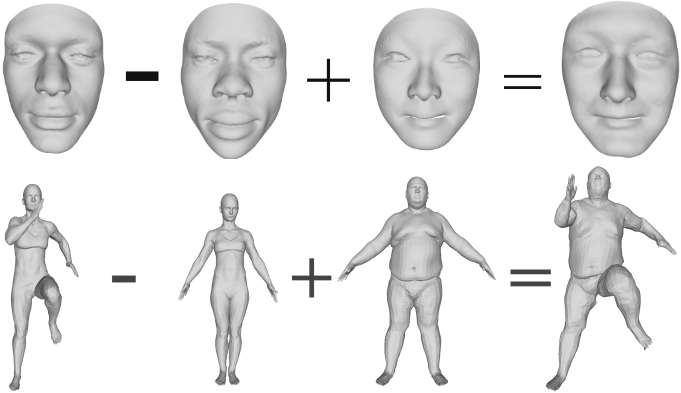}
   \caption{Analogies in MeIn3D and DFAUST}
\label{fig:analogies}
\end{figure}

\noindent\textbf{Shape Analogies} Fig \ref{fig:analogies}: We choose three meshes $A$, $B$, $C$, and construct a $D$ such that it satisfies $A$:$B$::$C$:$D$ using linear algebra in the latent space as in \cite{mikolov}: $e(B)-e(A)=e(D)-e(C)$ ($e(*)$ the encoding), where we then solve for $e(D)$ and decode it. This way we transfer a specific characteristic using meshes from our dataset.

\subsection{GAN evaluation}

\begin{figure}[h]
        \centering
        \includegraphics[width=0.7\columnwidth]{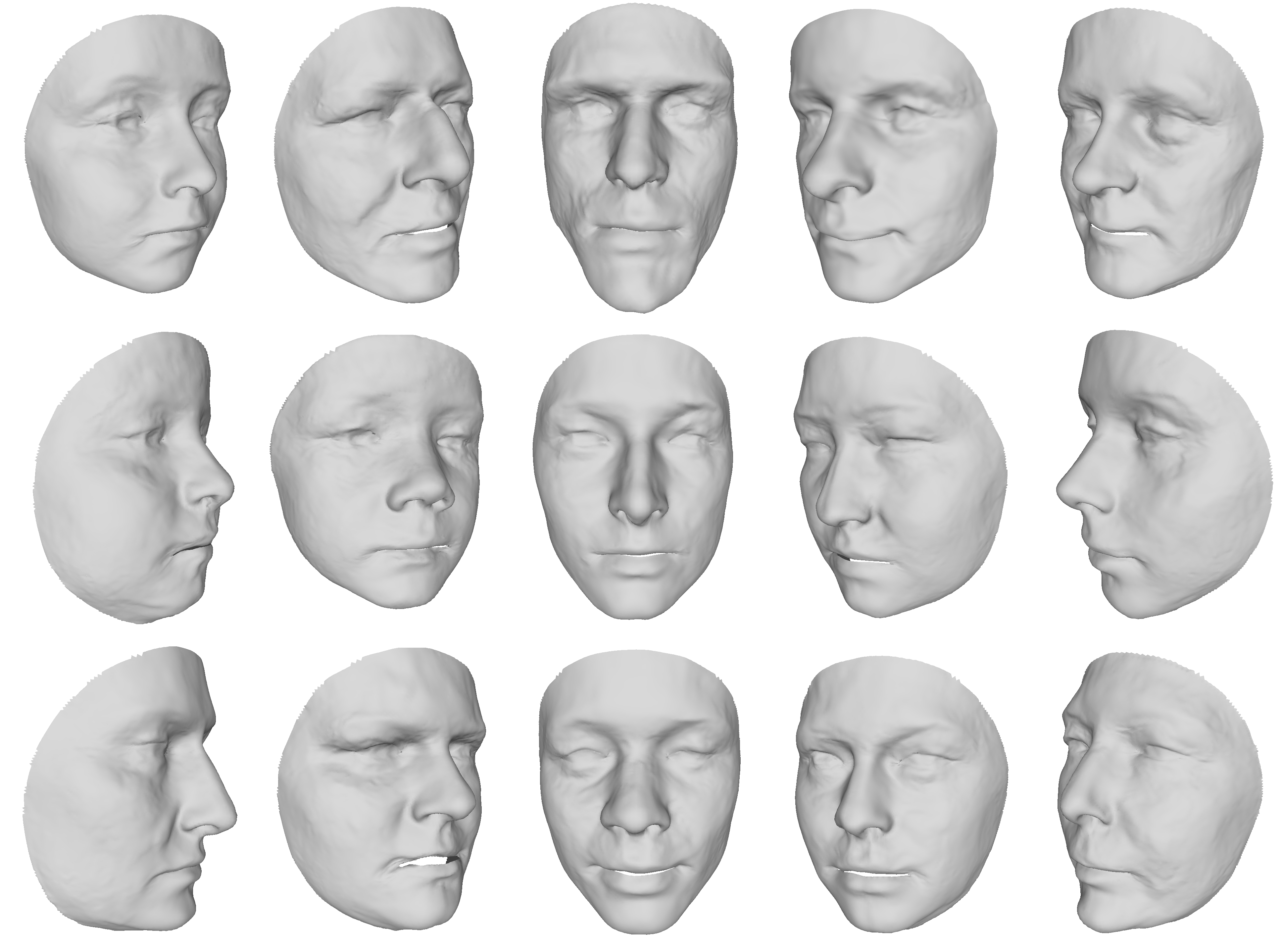}
        \caption{Generated identities from our intrinsic 3D GAN}
        \label{fig:gan_samples}
\end{figure}

In figure \ref{fig:gan_samples}, we sampled several faces from the latent distribution of the trained generator. Notice that they are realistically looking and, following the statistics of the dataset, span a large proportion of the real distribution of the human faces, in terms of ethnicity, gender and age.
Compared to the most popular approach for synthesizing faces, \ie the 3DMM, our model learns to produce fine details on the facial structure, making them hard to distinguish from real 3D scans, whereas the 3DMM, although it produces smooth surfaces, frequently makes it easy to tell the difference between real and artificially produced samples. We direct the reader to the supplementary material to compare with samples drawn from the 3DMM's latent space.
\section{Conclusion}
In this paper we introduced a representation learning and generative framework for fixed topology 3D deformable shapes, by using a mesh convolutional operator, spiral convolutions, that efficiently encodes the inductive bias of the fixed topology. We showcased the inherent representational power of the operator, as well as its reduced computational complexity, compared to prior work on graph convolutional operators and show that our mesh autoencoder achieves state-of-the-art results in mesh reconstruction. Finally, we present the generation capabilities of our models through vector space arithmetic, as well as by synthesising novel facial identities. Regarding future work, we plan to extend our framework to general graphs and 3D shapes of arbitrary topology, as well as to other domains that have capacity for an implicit ordering of their primitives, such as point clouds.
\section{Acknowledgements}
This research was partially supported by ERC Consolidator Grant No. 724228 (LEMAN), Google Research Faculty awards, and the Royal Society Wolfson Research Merit award. G. Bouritsas was funded by the Imperial College London, Department of Computing, PhD scholarship. Dr. Zafeiriou acknowledges support from a Google Faculty award and EPSRC fellowship Deform (EP/S010203/1). S. Ploumpis was suppored by EPSRC Project (EP/N007743/1) FACER2VM.

{\small
\bibliographystyle{ieee}
\bibliography{egbib}

\begin{thebibliography}{10}\itemsep=-1pt

\bibitem{achlioptas2017learning}
P.~Achlioptas, O.~Diamanti, I.~Mitliagkas, and L.~Guibas.
\newblock Learning representations and generative models for 3d point clouds.
\newblock {\em International Conference on Machine Learning (ICML)}, 2018.

\bibitem{pmlr-v70-arjovsky17a}
M.~Arjovsky, S.~Chintala, and L.~Bottou.
\newblock {W}asserstein generative adversarial networks.
\newblock In {\em Proceedings of the 34th International Conference on Machine
  Learning (ICML)}, 2017.

\bibitem{arsalan2017synthesizing}
A.~Arsalan~Soltani, H.~Huang, J.~Wu, T.~D. Kulkarni, and J.~B. Tenenbaum.
\newblock Synthesizing 3d shapes via modeling multi-view depth maps and
  silhouettes with deep generative networks.
\newblock In {\em Proceedings of the IEEE Conference on Computer Vision and
  Pattern Recognition (CVPR)}, 2017.

\bibitem{ben2018multi}
H.~Ben-Hamu, H.~Maron, I.~Kezurer, G.~Avineri, and Y.~Lipman.
\newblock Multi-chart generative surface modeling.
\newblock In {\em SIGGRAPH Asia 2018 Technical Papers}, page 215. ACM, 2018.

\bibitem{vetterMM}
V.~Blanz, T.~Vetter, et~al.
\newblock A morphable model for the synthesis of 3d faces.
\newblock In {\em SIGGRAPH}, 1999.

\bibitem{dfaust:CVPR:2017}
F.~Bogo, J.~Romero, G.~Pons-Moll, and M.~J. Black.
\newblock Dynamic {FAUST}: {R}egistering human bodies in motion.
\newblock In {\em Proceedings of the IEEE Conference on Computer Vision and
  Pattern Recognition (CVPR)}, 2017.

\bibitem{booth2018large}
J.~Booth, A.~Roussos, A.~Ponniah, D.~Dunaway, and S.~Zafeiriou.
\newblock Large scale 3d morphable models.
\newblock {\em International Journal of Computer Vision (IJCV)}, 2018.

\bibitem{booth20163d}
J.~Booth, A.~Roussos, S.~Zafeiriou, A.~Ponniah, and D.~Dunaway.
\newblock A 3d morphable model learnt from 10,000 faces.
\newblock In {\em Proceedings of the IEEE Conference on Computer Vision and
  Pattern Recognition (CVPR)}, 2016.

\bibitem{boscaini2016learning}
D.~Boscaini, J.~Masci, E.~Rodol{\`a}, and M.~Bronstein.
\newblock Learning shape correspondence with anisotropic convolutional neural
  networks.
\newblock In {\em Advances in Neural Information Processing Systems (NIPS)},
  2016.

\bibitem{bronstein2017geometric}
M.~M. Bronstein, J.~Bruna, Y.~LeCun, A.~Szlam, and P.~Vandergheynst.
\newblock Geometric deep learning: going beyond euclidean data.
\newblock {\em IEEE Signal Processing Magazine}, 34(4):18--42, 2017.

\bibitem{bruna2013spectral}
J.~Bruna, W.~Zaremba, A.~Szlam, and Y.~LeCun.
\newblock Spectral networks and locally connected networks on graphs.
\newblock In {\em International Conference on Learning Representations (ICLR)},
  2014.

\bibitem{cao2014facewarehouse}
C.~Cao, Y.~Weng, S.~Zhou, Y.~Tong, and K.~Zhou.
\newblock Facewarehouse: A 3d facial expression database for visual computing.
\newblock {\em IEEE Transactions on Visualization and Computer Graphics}, 2014.

\bibitem{chen2018learning}
Z.~Chen and H.~Zhang.
\newblock Learning implicit fields for generative shape modeling.
\newblock {\em CVPR}, 2019.

\bibitem{ELU}
D.~Clevert, T.~Unterthiner, and S.~Hochreiter.
\newblock Fast and accurate deep network learning by exponential linear units
  (elus).
\newblock {\em CoRR}, 2015.

\bibitem{defferrard2016convolutional}
M.~Defferrard, X.~Bresson, and P.~Vandergheynst.
\newblock Convolutional neural networks on graphs with fast localized spectral
  filtering.
\newblock In {\em Advances in Neural Information Processing Systems (NIPS)},
  2016.

\bibitem{chebnet}
M.~Defferrard, X.~Bresson, and P.~Vandergheynst.
\newblock Convolutional neural networks on graphs with fast localized spectral
  filtering.
\newblock {\em Advances in Neural Information Processing Systems (NIPS)}, 2016.

\bibitem{duvenaud2015convolutional}
D.~K. Duvenaud, D.~Maclaurin, J.~Iparraguirre, R.~Bombarell, T.~Hirzel,
  A.~Aspuru-Guzik, and R.~P. Adams.
\newblock Convolutional networks on graphs for learning molecular fingerprints.
\newblock In {\em Advances in Neural Information Processing systems (NIPS)},
  2015.

\bibitem{fey2018splinecnn}
M.~Fey, J.~E. Lenssen, F.~Weichert, and H.~M{\"u}ller.
\newblock Splinecnn: Fast geometric deep learning with continuous b-spline
  kernels.
\newblock In {\em Proceedings of the IEEE Conference on Computer Vision and
  Pattern Recognition (CVPR)}, 2018.

\bibitem{gilmer2017neural}
J.~Gilmer, S.~S. Schoenholz, P.~F. Riley, O.~Vinyals, and G.~E. Dahl.
\newblock Neural message passing for quantum chemistry.
\newblock In {\em Proceedings of the 34th International Conference on Machine
  Learning (ICML)}, 2017.

\bibitem{girdhar2016learning}
R.~Girdhar, D.~F. Fouhey, M.~Rodriguez, and A.~Gupta.
\newblock Learning a predictable and generative vector representation for
  objects.
\newblock In {\em European Conference on Computer Vision (ECCV)}. Springer,
  2016.

\bibitem{gulrajani2017improved}
I.~Gulrajani, F.~Ahmed, M.~Arjovsky, V.~Dumoulin, and A.~C. Courville.
\newblock Improved training of wasserstein gans.
\newblock In {\em Advances in Neural Information Processing Systems (NIPS)},
  2017.

\bibitem{kipfGCN}
T.~N. Kipf and M.~Welling.
\newblock Semi-supervised classification with graph convolutional networks.
\newblock In {\em International Conference on Learning Representations (ICLR)},
  2017.

\bibitem{kolotouros2019convolutional}
N.~Kolotouros, G.~Pavlakos, and K.~Daniilidis.
\newblock Convolutional mesh regression for single-image human shape
  reconstruction.
\newblock In {\em CVPR}, 2019.

\bibitem{FLAME:SiggraphAsia2017}
T.~Li, T.~Bolkart, M.~J. Black, H.~Li, and J.~Romero.
\newblock Learning a model of facial shape and expression from {4D} scans.
\newblock {\em ACM Transactions on Graphics, (Proc. SIGGRAPH Asia)}, 2017.

\bibitem{spirals}
I.~Lim, A.~Dielen, M.~Campen, and L.~Kobbelt.
\newblock A simple approach to intrinsic correspondence learning on
  unstructured 3d meshes.
\newblock {\em Proceedings of the European Conference on Computer Vision
  Workshops (ECCVW)}, 2018.

\bibitem{litany2018deformable}
O.~Litany, A.~Bronstein, M.~Bronstein, and A.~Makadia.
\newblock Deformable shape completion with graph convolutional autoencoders.
\newblock In {\em Proceedings of the IEEE Conference on Computer Vision and
  Pattern Recognition (CVPR)}, pages 1886--1895, 2018.

\bibitem{SMPL:2015}
M.~Loper, N.~Mahmood, J.~Romero, G.~Pons-Moll, and M.~J. Black.
\newblock {SMPL}: A skinned multi-person linear model.
\newblock {\em ACM Trans. Graphics (Proc. SIGGRAPH Asia)}, 34(6):248:1--248:16,
  Oct. 2015.

\bibitem{masci2015geodesic}
J.~Masci, D.~Boscaini, M.~Bronstein, and P.~Vandergheynst.
\newblock Geodesic convolutional neural networks on riemannian manifolds.
\newblock In {\em Proceedings of the IEEE International Conference on Computer
  Vision Workshops (ICCVW)}, pages 37--45, 2015.

\bibitem{maturana2015voxnet}
D.~Maturana and S.~Scherer.
\newblock Voxnet: A 3d convolutional neural network for real-time object
  recognition.
\newblock In {\em 2015 IEEE/RSJ International Conference on Intelligent Robots
  and Systems (IROS)}. IEEE, 2015.

\bibitem{mescheder2018occupancy}
L.~Mescheder, M.~Oechsle, M.~Niemeyer, S.~Nowozin, and A.~Geiger.
\newblock Occupancy networks: Learning 3d reconstruction in function space.
\newblock {\em CVPR}, 2019.

\bibitem{mikolov}
T.~Mikolov, I.~Sutskever, K.~Chen, G.~S. Corrado, and J.~Dean.
\newblock Distributed representations of words and phrases and their
  compositionality.
\newblock In {\em Advances in Neural Information Processing Systems (NIPS)}.
  2013.

\bibitem{monti2017geometric}
F.~Monti, D.~Boscaini, J.~Masci, E.~Rodola, J.~Svoboda, and M.~M. Bronstein.
\newblock Geometric deep learning on graphs and manifolds using mixture model
  cnns.
\newblock In {\em Proceedings of the IEEE Conference on Computer Vision and
  Pattern Recognition (CVPR)}, 2017.

\bibitem{moschoglou20193dfacegan}
S.~Moschoglou, S.~Ploumpis, M.~Nicolaou, and S.~Zafeiriou.
\newblock 3dfacegan: Adversarial nets for 3d face representation, generation,
  and translation.
\newblock {\em arXiv preprint arXiv:1905.00307}, 2019.

\bibitem{park2019deepsdf}
J.~J. Park, P.~Florence, J.~Straub, R.~Newcombe, and S.~Lovegrove.
\newblock Deepsdf: Learning continuous signed distance functions for shape
  representation.
\newblock {\em CVPR}, 2019.

\bibitem{ploumpis2019combining}
S.~Ploumpis, H.~Wang, N.~Pears, W.~A. Smith, and S.~Zafeiriou.
\newblock Combining 3d morphable models: A large scale face-and-head model.
\newblock {\em CVPR}, 2019.

\bibitem{qi2017pointnet}
C.~R. Qi, H.~Su, K.~Mo, and L.~J. Guibas.
\newblock Pointnet: Deep learning on point sets for 3d classification and
  segmentation.
\newblock In {\em Proceedings of the IEEE Conference on Computer Vision and
  Pattern Recognition (CVPR)}, 2017.

\bibitem{qi2016volumetric}
C.~R. Qi, H.~Su, M.~Nie{\ss}ner, A.~Dai, M.~Yan, and L.~J. Guibas.
\newblock Volumetric and multi-view cnns for object classification on 3d data.
\newblock In {\em Proceedings of the IEEE Conference on Computer Vision and
  Pattern Recognition (CVPR)}, 2016.

\bibitem{qi2017pointnet++}
C.~R. Qi, L.~Yi, H.~Su, and L.~J. Guibas.
\newblock Pointnet++: Deep hierarchical feature learning on point sets in a
  metric space.
\newblock In {\em Advances in Neural Information Processing Systems (NIPS)},
  pages 5099--5108, 2017.

\bibitem{coma}
A.~Ranjan, T.~Bolkart, S.~Sanyal, and M.~J. Black.
\newblock Generating 3d faces using convolutional mesh autoencoders.
\newblock {\em Proceedings of the European Conference on Computer Vision
  (ECCV)}, 2018.

\bibitem{MANO:SIGGRAPHASIA:2017}
J.~Romero, D.~Tzionas, and M.~J. Black.
\newblock Embodied hands: Modeling and capturing hands and bodies together.
\newblock {\em ACM Transactions on Graphics, (Proc. SIGGRAPH Asia)}, 2017.

\bibitem{sharma2016vconv}
A.~Sharma, O.~Grau, and M.~Fritz.
\newblock Vconv-dae: Deep volumetric shape learning without object labels.
\newblock In {\em European Conference on Computer Vision}, pages 236--250.
  Springer, 2016.

\bibitem{velivckovic2017graph}
P.~Veli{\v{c}}kovi{\'c}, G.~Cucurull, A.~Casanova, A.~Romero, P.~Li{\`o}, and
  Y.~Bengio.
\newblock Graph attention networks.
\newblock In {\em International Conference on Learning Representations (ICLR)},
  2018.

\bibitem{verma2018feastnet}
N.~Verma, E.~Boyer, and J.~Verbeek.
\newblock Feastnet: Feature-steered graph convolutions for 3d shape analysis.
\newblock In {\em Proceedings of the IEEE Conference on Computer Vision and
  Pattern Recognition(CVPR)}, pages 2598--2606, 2018.

\bibitem{wang2018pixel2mesh}
N.~Wang, Y.~Zhang, Z.~Li, Y.~Fu, W.~Liu, and Y.-G. Jiang.
\newblock Pixel2mesh: Generating 3d mesh models from single rgb images.
\newblock In {\em ECCV}, 2018.

\bibitem{wu2016learning}
J.~Wu, C.~Zhang, T.~Xue, B.~Freeman, and J.~Tenenbaum.
\newblock Learning a probabilistic latent space of object shapes via 3d
  generative-adversarial modeling.
\newblock In {\em Advances in Neural Information Processing Systems (NIPS)},
  2016.

\bibitem{wu20153d}
Z.~Wu, S.~Song, A.~Khosla, F.~Yu, L.~Zhang, X.~Tang, and J.~Xiao.
\newblock 3d shapenets: A deep representation for volumetric shapes.
\newblock In {\em Proceedings of the IEEE conference on Computer Vision and
  Pattern Recognition (CVPR)}, 2015.

\bibitem{yang2018foldingnet}
Y.~Yang, C.~Feng, Y.~Shen, and D.~Tian.
\newblock Foldingnet: Point cloud auto-encoder via deep grid deformation.
\newblock In {\em Proceedings of the IEEE Conference on Computer Vision and
  Pattern Recognition (CVPR)}, 2018.

\bibitem{yi2017syncspeccnn}
L.~Yi, H.~Su, X.~Guo, and L.~J. Guibas.
\newblock Syncspeccnn: Synchronized spectral cnn for 3d shape segmentation.
\newblock In {\em Proceedings of the IEEE Conference on Computer Vision and
  Pattern Recognition (CVPR)}, 2017.

\bibitem{ying2018hierarchical}
Z.~Ying, J.~You, C.~Morris, X.~Ren, W.~Hamilton, and J.~Leskovec.
\newblock Hierarchical graph representation learning with differentiable
  pooling.
\newblock In {\em Advances in Neural Information Processing Systems (NIPS)},
  2018.

\bibitem{yu2015multi}
F.~Yu and V.~Koltun.
\newblock Multi-scale context aggregation by dilated convolutions.
\newblock {\em International Conference on Learning Representations (ICLR)},
  2016.

\end{thebibliography}
}

\end{document}